\documentclass{article}

\usepackage{cig_preprint,times}

\usepackage{amsmath,amsfonts,bm}

\newcommand{\TableOneRows}{%
\multicolumn{9}{l}{\textit{Published HarnessFix results (3-run means)}} \\
\addlinespace[2pt]
Initial Harness~\citep{chen2026harnessfix} & 45.3 &  & 36.7 &  & 17.6 &  & 33.2 &  \\
GEPA~\citep{agrawal2026gepa} & 46.7 &  & 37.4 &  & 19.6 &  & 34.6 &  \\
SCOPE~\citep{pei2025scope} & 48.3 &  & 38.9 &  & 20.6 &  & 35.9 &  \\
ReCreate~\citep{hao2026recreate} & 51.7 &  & 39.3 &  & 21.6 &  & 37.5 &  \\
Meta-Harness~\citep{lee2026metaharness} & 54.7 &  & 40.4 &  & 23.5 &  & 39.5 &  \\
HarnessFix~\citep{chen2026harnessfix} & \textbf{57.3} &  & \textbf{43.0} &  & \textbf{26.5} &  & \textbf{42.3} &  \\
\midrule
\multicolumn{9}{l}{\textit{Our runs (3-run means)}} \\
\addlinespace[2pt]
Reproduced Initial Harness & 45.7 &  & 35.6 &  & 17.6 &  & 33.0 &  \\
Full History (our implementation) & 56.7 &  & 40.0 &  & 23.5 &  & 40.1 &  \\
CIG (Ours) & \textbf{60.0} & {\scriptsize \textcolor{gaincolor}{(+14.3)}} & \textbf{44.4} & {\scriptsize \textcolor{gaincolor}{(+8.8)}} & \textbf{29.4} & {\scriptsize \textcolor{gaincolor}{(+11.8)}} & \textbf{44.6} & {\scriptsize \textcolor{gaincolor}{(+11.6)}} \\
}

\newcommand{\TableTwoRows}{%
Qwen3.5-Plus & 51.5 & \textbf{60.6} & 59.7 & \textcolor{gaincolor}{+9.1} & \textcolor{gaincolor}{+8.2} & 7.1 & 0.0 & 0.0 \\
Qwen3.7-Flash & 41.3 & \textbf{53.8} & 49.4 & \textcolor{gaincolor}{+12.5} & \textcolor{gaincolor}{+8.2} & 0.0 & 0.0 & 2.0 \\
DeepSeek-v4-Flash & 51.3 & \textbf{55.2} & 51.7 & \textcolor{gaincolor}{+3.9} & \textcolor{gaincolor}{+0.4} & 0.0 & 0.0 & 3.9 \\
GLM-5.2-Fast-Preview & 50.4 & \textbf{54.7} & 50.3 & \textcolor{gaincolor}{+4.3} & \textcolor{losscolor}{-0.1} & 0.0 & 0.0 & 2.5 \\
MiniMax-M2.5 & 45.4 & \textbf{48.9} & 47.8 & \textcolor{gaincolor}{+3.5} & \textcolor{gaincolor}{+2.4} & 0.0 & 0.2 & 0.1 \\
}

\newcommand{\TableThreeRows}{%
GPT-5.5 & \mbox{58.6{\scriptsize\(\,\pm\,0.5\)}} & \mbox{\textbf{61.0}{\scriptsize\(\,\pm\,0.0\)}} & \textcolor{gaincolor}{+2.4} \\
GPT-5.6 Sol & \mbox{59.0{\scriptsize\(\,\pm\,1.0\)}} & \mbox{\textbf{64.0}{\scriptsize\(\,\pm\,1.0\)}} & \textcolor{gaincolor}{+5.0} \\
GLM-5.3 & \mbox{58.2{\scriptsize\(\,\pm\,1.1\)}} & \mbox{\textbf{65.6}{\scriptsize\(\,\pm\,0.5\)}} & \textcolor{gaincolor}{+7.4} \\
Kimi K3 & \mbox{56.6{\scriptsize\(\,\pm\,0.5\)}} & \mbox{\textbf{65.4}{\scriptsize\(\,\pm\,0.5\)}} & \textcolor{gaincolor}{+8.8} \\
DeepSeek-V4-Pro & \mbox{59.0{\scriptsize\(\,\pm\,0.7\)}} & \mbox{\textbf{63.0}{\scriptsize\(\,\pm\,1.0\)}} & \textcolor{gaincolor}{+4.0} \\
}

\newcommand{\TableFourRows}{%
\textbf{\mbox{M0 (Full History)}} & Raw & None & Reset & Indep. & 58.0 &  & 58.0 & 56.4 &  & 56.8 \\
\textbf{\mbox{M1 (+ Typed state)}} & Typed & None & Reset & Indep. & 59.6 & {\scriptsize \textcolor{gaincolor}{(+1.6)}} & 59.1 & 57.4 & {\scriptsize \textcolor{gaincolor}{(+1.0)}} & 57.9 \\
\textbf{\mbox{M2 (+ Relations)}} & Typed & Correct & Reset & Indep. & 60.8 & {\scriptsize \textcolor{gaincolor}{(+1.2)}} & 60.5 & 58.8 & {\scriptsize \textcolor{gaincolor}{(+1.4)}} & 59.2 \\
\rowcolor{diagnosticbg} \mbox{$\hookrightarrow$\,\textit{B1 (Shuffled)}} & Typed & Shuffled & Reset & Indep. & 59.0 & {\scriptsize \textcolor{losscolor}{(-1.8)}} & 59.2 & 57.4 & {\scriptsize \textcolor{losscolor}{(-1.4)}} & 58.1 \\
\textbf{\mbox{M3 (+ Revision)}} & Typed & Correct & Revise & Indep. & 62.2 & {\scriptsize \textcolor{gaincolor}{(+1.4)}} & 61.8 & 59.6 & {\scriptsize \textcolor{gaincolor}{(+0.8)}} & 60.4 \\
\rowcolor{diagnosticbg} \mbox{$\hookrightarrow$\,\textit{B2 (Append-only)}} & Typed & Correct & Append-only & Indep. & 61.0 & {\scriptsize \textcolor{losscolor}{(-1.2)}} & 60.6 & 58.8 & {\scriptsize \textcolor{losscolor}{(-0.8)}} & 59.6 \\
\textbf{\mbox{M4 (Full CIG)}} & Typed & Correct & Revise & Graph & \textbf{63.6} & {\scriptsize \textcolor{gaincolor}{(+1.4)}} & \textbf{63.0} & \textbf{61.4} & {\scriptsize \textcolor{gaincolor}{(+1.8)}} & \textbf{61.5} \\
\rowcolor{diagnosticbg} \mbox{$\hookrightarrow$\,\textit{B3 (Local only)}} & Typed & Local only & Revise & Records & 62.0 & {\scriptsize \textcolor{losscolor}{(-1.6)}} & 61.0 & 59.2 & {\scriptsize \textcolor{losscolor}{(-2.2)}} & 59.4 \\
}

\newcommand{\TableFourVarianceRows}{%
M0 (Full History) & \mbox{58.0{\scriptsize\(\,\pm\,0.0\)}} & \mbox{56.4{\scriptsize\(\,\pm\,0.9\)}} \\
M1 (+ Typed State) & \mbox{59.6{\scriptsize\(\,\pm\,0.5\)}} & \mbox{57.4{\scriptsize\(\,\pm\,0.5\)}} \\
M2 (+ Logic Relations) & \mbox{60.8{\scriptsize\(\,\pm\,0.8\)}} & \mbox{58.8{\scriptsize\(\,\pm\,0.4\)}} \\
\rowcolor{diagnosticbg} \textit{B1 (Shuffled Relations)} & \mbox{59.0{\scriptsize\(\,\pm\,0.0\)}} & \mbox{57.4{\scriptsize\(\,\pm\,0.5\)}} \\
M3 (+ Revision-aware Persistence) & \mbox{62.2{\scriptsize\(\,\pm\,0.8\)}} & \mbox{59.6{\scriptsize\(\,\pm\,0.9\)}} \\
\rowcolor{diagnosticbg} \textit{B2 (Append-only Persistence)} & \mbox{61.0{\scriptsize\(\,\pm\,0.0\)}} & \mbox{58.8{\scriptsize\(\,\pm\,0.4\)}} \\
M4 (Full CIG) & \mbox{63.6{\scriptsize\(\,\pm\,1.1\)}} & \mbox{61.4{\scriptsize\(\,\pm\,0.9\)}} \\
\rowcolor{diagnosticbg} \textit{B3 (w/o Cross-record Relations)} & \mbox{62.0{\scriptsize\(\,\pm\,0.0\)}} & \mbox{59.2{\scriptsize\(\,\pm\,0.8\)}} \\
}

\usepackage{xcolor}
\usepackage{colortbl}
\usepackage{booktabs}
\usepackage{amssymb}
\usepackage{graphicx}
\usepackage{placeins}
\usepackage{tabularx}
\usepackage{algorithm}
\usepackage{algpseudocode}
\usepackage{wrapfig}
\usepackage{ragged2e}
\usepackage{capt-of}
\usepackage{xspace}
\usepackage[most]{tcolorbox}
\usepackage{needspace}
\usepackage{enumitem}
\usepackage{titletoc}
\definecolor{darkblue}{HTML}{174A75}
\definecolor{featuregreen}{HTML}{176647}
\definecolor{gaincolor}{HTML}{287A38}
\definecolor{losscolor}{HTML}{B83332}
\definecolor{diagnosticbg}{HTML}{EEF3FA}
\newcommand{\featureyes}{\textcolor{featuregreen}{$\checkmark$}}
\newcommand{\featureunspecified}{\textcolor{gray}{--}}
\definecolor{mycitecolor}{HTML}{2B6EA6}
\definecolor{mylinkcolor}{HTML}{E74D3B}
\definecolor{myblue}{HTML}{4594c1}
\definecolor{myurlcolor}{HTML}{980000}
\definecolor{evidencebg}{HTML}{DDEAF2}
\definecolor{hypothesisbg}{HTML}{F2E9D2}
\definecolor{interventionbg}{HTML}{F1DEDF}
\definecolor{outcomebg}{HTML}{DDEBDD}
\definecolor{evidencebar}{HTML}{7391A3}
\definecolor{hypothesisbar}{HTML}{9C8A5F}
\definecolor{interventionbar}{HTML}{A27C80}
\definecolor{outcomebar}{HTML}{78917A}
\usepackage[colorlinks=true,linkcolor=mylinkcolor,citecolor=mycitecolor,urlcolor=myurlcolor]{hyperref}

\usepackage{url}
\newcommand{\concepttag}[2]{%
  \begingroup\setlength{\fboxsep}{1.2pt}%
  \colorbox{#1}{\strut\textsc{#2}}%
  \endgroup}

\newcommand{\evidencetag}{\concepttag{evidencebg}{Evidence}\xspace}
\newcommand{\hypothesistag}{\concepttag{hypothesisbg}{Hypothesis}\xspace}
\newcommand{\interventiontag}{\concepttag{interventionbg}{Intervention}\xspace}
\newcommand{\outcometag}{\concepttag{outcomebg}{Outcome}\xspace}

\newcommand{\evidenceterm}{\textsc{Evidence}\xspace}
\newcommand{\hypothesisterm}{\textsc{Hypothesis}\xspace}
\newcommand{\interventionterm}{\textsc{Intervention}\xspace}
\newcommand{\outcometerm}{\textsc{Outcome}\xspace}

\newcommand{\rqmark}[1]{\texorpdfstring{\textcolor{darkblue}{\textbf{\textup{RQ#1}}}}{RQ#1}}
\newtcolorbox{researchquestion}[1]{
  colback=black!3, colframe=black!18, boxrule=0.4pt, arc=1pt,
  left=6pt, right=6pt, top=4pt, bottom=4pt,
  before skip=6pt, after skip=6pt, fontupper=\small,
  before upper={\textbf{#1}\enspace}}
\newtcolorbox{casenodebox}[3]{
  enhanced,
  breakable,
  colback=#1!72!white,
  colframe=#2,
  colbacktitle=#2,
  coltitle=white,
  fonttitle=\bfseries,
  title={#3},
  boxrule=0.55pt,
  arc=2pt,
  outer arc=2pt,
  left=7pt,
  right=7pt,
  top=5pt,
  bottom=6pt,
  toptitle=2pt,
  bottomtitle=2pt,
  before skip=7pt,
  after skip=7pt
}

\newcolumntype{Y}{>{\raggedright\arraybackslash}X}
\newsavebox{\alignedtext}

\title{Causal Improvement Graph for Agentic Harness Optimization}

\author{\parbox[t]{\textwidth}{\small\raggedright
\textbf{Junjie Zhang\textsuperscript{1}},
\textbf{Shunyu Liu\textsuperscript{1}},
\textbf{Haoyu Wang\textsuperscript{1}},
\textbf{Ting-En Lin\textsuperscript{2}},
\textbf{Yongbin Li\textsuperscript{2}},
\textbf{Dacheng Tao\textsuperscript{1}}\\[6pt]
\normalfont\small\textsuperscript{1}Generative AI Lab, College of Computing and Data Science, Nanyang Technological University,\\
Singapore 639798\\
\textsuperscript{2}Tongyi Lab, Alibaba Group
}
}
\iclrfinalcopy
\hypersetup{pdftitle={Causal Improvement Graph for Agentic Harness Optimization},
pdfauthor={Junjie Zhang, Shunyu Liu, Haoyu Wang, Ting-En Lin, Yongbin Li, Dacheng Tao}}

\begin{document}

\maketitle
\pagestyle{plain}
\thispagestyle{plain}

\vspace{-18pt}
\begin{abstract}
Agentic Harness is the runtime that constructs task context and controls
execution flow, thereby shaping overall agent performance. Given a fixed model and external
evaluation, automated Harness optimization seeks
to improve this runtime through an iterative proposal--evaluation loop to better solve target tasks. Existing meta-harness methods mainly adopt \emph{proposer-centric discovery}, in which an LLM-based proposer integrates accumulated experimental findings to determine subsequent Harness
revisions. This places the burden of maintaining the evolving improvement state on the proposer as history expands and its underlying experimental logic becomes harder to discern. In this paper, we
introduce the \emph{Causal
Improvement Graph} (CIG), a graph-governed meta-harness framework that externalizes the evolving improvement state in a persistent
graph, allowing prior findings to directly govern subsequent Harness optimization through local proposer operations. CIG grows and links
\evidenceterm, \hypothesisterm, \interventionterm, and
\outcometerm nodes to represent what was observed, how it may be explained, how to test that
explanation, and what the evaluation reveals. Their structural relations preserve how the improvement state changes across iterations, allowing local proposers to build directly on relations among prior findings rather than recover them from raw history. Across various agent tasks, CIG discovers stronger Harnesses than previous meta-harness baselines and remains robust to the choice of
task solver and proposer. Structural ablations further support the design of an explicit
improvement state with graph-governed evolution.
\end{abstract}

\vspace{-8pt}
\section{Introduction}
\vspace{-3pt}

As foundation models increasingly operate as agents, Harness design has emerged alongside model
development as a major route to improving agent performance~\citep{lee2026metaharness,lin2026ahe}.
A \emph{Harness} is the programmable runtime that turns model capabilities into task behavior by
constructing context, mediating tools, and coordinating execution, as realized
in agent systems such as Codex, Claude Code, and OpenClaw~\citep{lopopolo2026harness,anthropic2026dynamicworkflows,openclaw2026}.
While model optimization has a mature training stack~\citep{brown2020gpt3,ouyang2022instructgpt},
Harness optimization remains substantially dependent on engineering judgments about context,
tools, and execution~\citep{lee2026metaharness,lin2026ahe}. Making the observations and reasoning
behind these judgments reusable is a central challenge for automated Harness optimization.

Automated Harness optimization makes the harness an object of the search process~\citep{lou2026autoharness,luo2026autodesign}.
Meta-Harness~\citep{lee2026metaharness} enables end-to-end search for stronger Harnesses
through an agentic proposer that autonomously inspects prior code, scores, and execution traces
via the filesystem, with minimal prescribed structure in the outer loop.
More structured approaches~\citep{lin2026ahe,chen2026harnessfix,zhang2026selfharness,chen2026harnessx}
organize execution feedback into diagnostic evidence, explicit change proposals, and validation
procedures to make Harness updates more targeted and verifiable.
Other approaches~\citep{lee2026rhi,zhang2026darwinx} organize reuse across rounds through
pairwise feedback over revision histories and archives of alternative Harness lineages for
selection and recombination.
These approaches accumulate increasingly rich histories, candidate archives, and diagnostic records.
As these materials grow substantially, the relationships among prior findings and their implications for further
optimization can become harder to track~\citep{park2026autosaddler}, leaving the proposer to reconstruct the evolving improvement state.
\newpage
\begingroup
\begin{wrapfigure}{r}{0.63\textwidth}
    \vspace{-2pt}
    \setlength{\abovecaptionskip}{0pt}
    \setlength{\belowcaptionskip}{0pt}
    \centering
    \includegraphics[width=\linewidth]{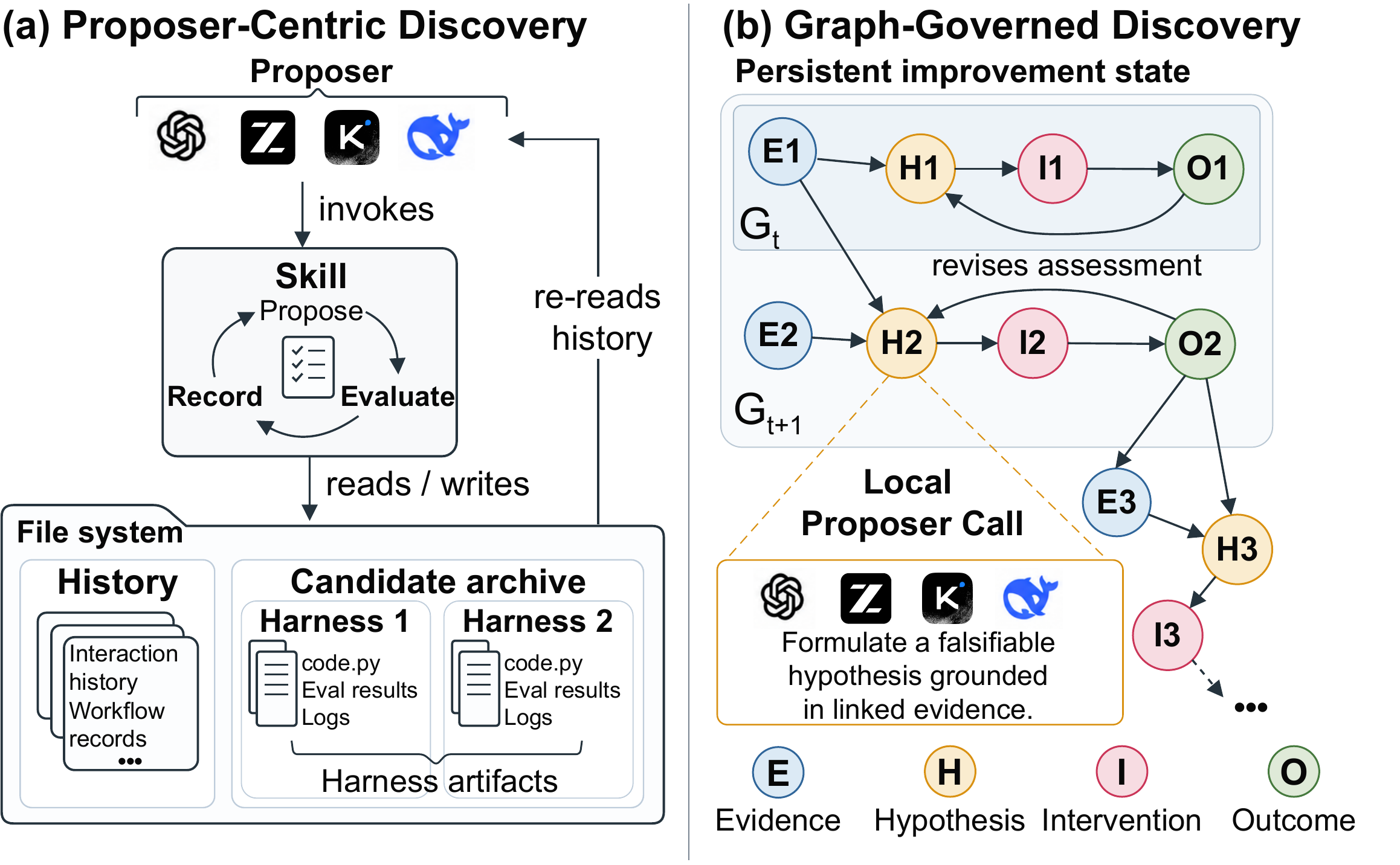}\par
    \vspace{-2pt}
    \caption{Comparison of proposer-centric discovery and graph-governed discovery. Left: a proposer
    follows a prescribed procedure to manage an end-to-end process over interaction records and candidate
    artifacts. Right: CIG makes the growing graph the persistent improvement state, invokes
    proposers for locally scoped operations, and updates the graph and its assessments using
    externally evaluated outcomes.}
    \label{fig:candidate-vs-graph}
    \vspace{-20pt}
\end{wrapfigure}

This reconstruction burden can reduce proposal efficiency and leave subsequent optimization
insufficiently guided by accumulated experience and prior progress.
To support further discovery, history should grow into an explicit, revisable improvement state
that preserves accumulated experience, maintains an evolving understanding of that experience,
and guides subsequent Harness optimization.

In this paper, we introduce the \textbf{Causal Improvement Graph (CIG)}, which shifts Harness
optimization from proposer-centric discovery to graph-governed discovery. By
linking \evidencetag, \hypothesistag, \interventiontag, and \outcometag nodes, CIG organizes raw
improvement history into a persistent improvement state that connects observations, explanations,
proposed changes, and evaluated outcomes to guide further Harness optimization. As illustrated in
Figure~\ref{fig:candidate-vs-graph}, CIG maintains the graph as a persistent improvement state
and invokes proposers for locally scoped operations. Externally evaluated outcomes revise
hypothesis assessments and update the state that guides subsequent Harness optimization.

The main contributions of this work are:
\par
\endgroup

\vspace{-4pt}
\begin{itemize}[leftmargin=*,labelsep=0.5em,itemsep=3pt,topsep=0pt,partopsep=0pt,parsep=0pt]
    \item \textbf{Graph-Governed Discovery.} We identify the burden of reconstructing evolving
    improvement state in proposer-centric discovery and reformulate Harness optimization as
    graph-governed discovery, guided by a persistent improvement state.
    \item \textbf{Causal Improvement Graph.} We introduce CIG, which grows, links, and updates
    \evidenceterm, \hypothesisterm, \interventionterm, and \outcometerm nodes to represent
    the improvement state and guide subsequent Harness optimization.
    \item \textbf{Empirical effectiveness and structural attribution.} We demonstrate CIG's
    effectiveness in discovering stronger Harnesses across tasks and its robustness to the
    choice of solver and proposer. Ablation studies
    clarify how CIG's structural design contributes to these gains.
\end{itemize}

\begin{figure}[t]
    \centering
    \includegraphics[width=\linewidth]{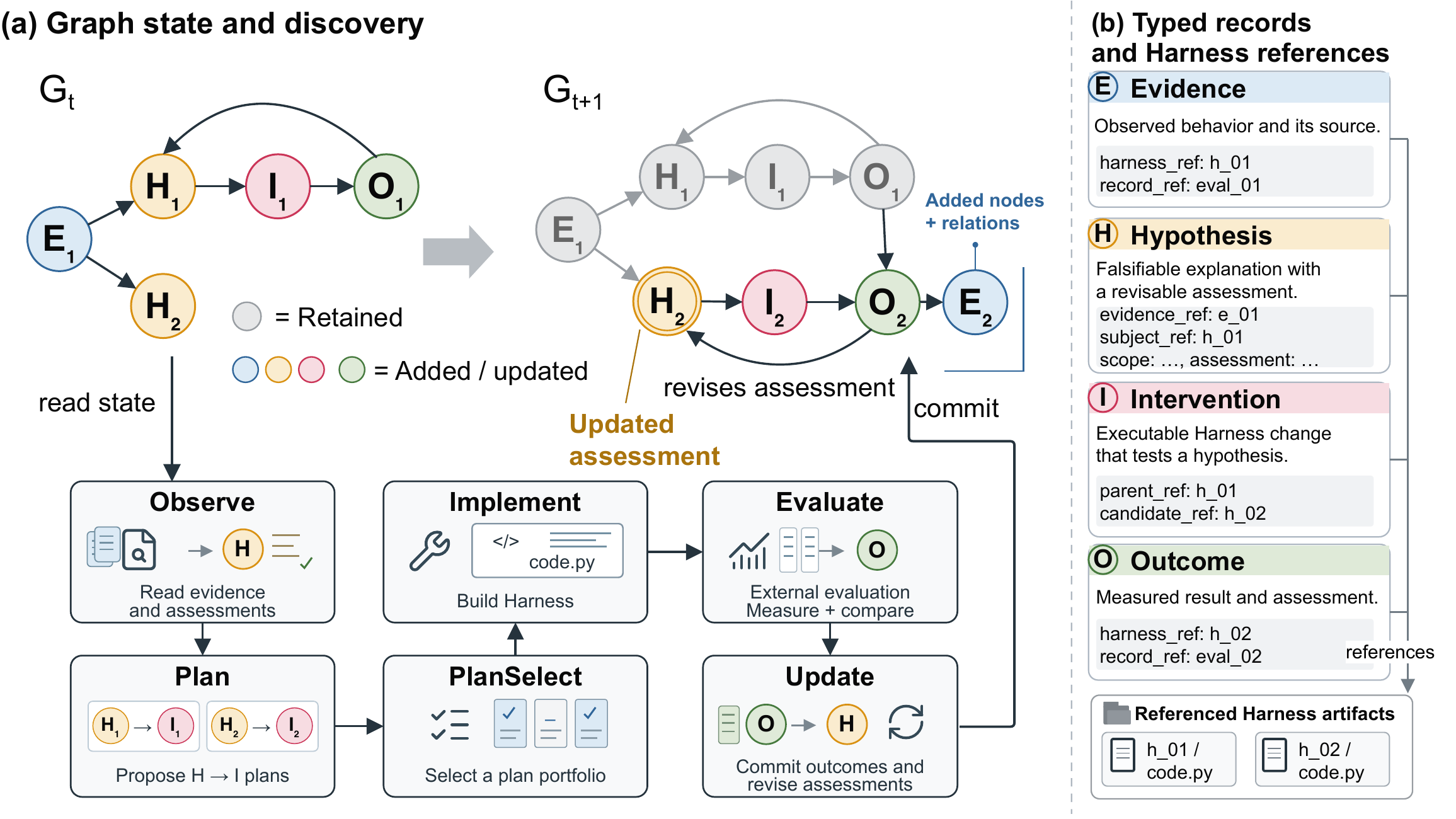}
    \caption{Improvement state transitions in CIG. Left: graph-governed discovery operations transform
    $G_t$ into $G_{t+1}$ by adding and linking evidence, hypothesis, intervention, and outcome nodes
    and revising existing hypothesis assessments based on evaluated outcomes.
    Right: typed records represent the improvement state and reference the corresponding Harness artifacts.}
    \label{fig:cig-overview}
    \vspace{-4pt}
\end{figure}

\vspace{-8pt}
\section{Related Work}

\vspace{-6pt}
\paragraph{Harness evolution.}
Agent design and workflow search provide foundations for Harness evolution: ADAS~\citep{hu2024adas}
generates programs from archived designs, while GPTSwarm~\citep{zhuge2024gptswarm} and
AFlow~\citep{zhang2024aflow} organize optimization around computational graphs and
code-represented workflows, respectively.
At the Harness level, Meta-Harness~\citep{lee2026metaharness} searches executable code using
prior candidates, scores, and traces.
To make these updates more explicitly grounded in execution evidence, AHE~\citep{lin2026ahe}
structures that evidence and verifies modification predictions against subsequent outcomes,
while HarnessFix~\citep{chen2026harnessfix} connects structured trace diagnosis to targeted
repairs and retains records of accepted and rejected changes.
\textbf{CIG} makes the evolving relations among evidence, hypotheses, interventions, and outcomes
an operational state: hypotheses retain revisable assessments across experiments, and their
current relations and assessments guide subsequent planning and portfolio selection.
Recent work~\citep{zhou2026hsi,park2026autosaddler,du2026livingharness,ren2026evigraph}
further explores explicit evolution procedures and persistent state for Harness evolution,
procedural adaptation, and autonomous research.
Appendix~\ref{app:related-comparison} discusses these methods in greater detail, comparing their
state representations and update targets with particular attention to graph-based approaches.

\vspace{-6pt}
\paragraph{Self-improving agents.}
Self-improving agents~\citep{zelikman2024stop,zhang2025dgm} use feedback to revise task-solving
behavior and, increasingly, the procedures that produce these revisions.
RHI~\citep{lee2026rhi} refines prompt-level Harness specifications using pairwise execution
feedback, while Hyperagents~\citep{zhang2026hyperagents} makes the meta-level modification
program editable and MetaSkill-Evolve~\citep{wang2026metaskillevolve} jointly evolves task
skills and the strategies used to improve them.
\textbf{CIG} focuses on the improvement state supporting this process: evaluated Harness
changes revise linked hypotheses and inform which interventions to attempt next.

\vspace{-6pt}
\section{Method}
\label{sec:method}

In this section, we formalize the Harness optimization problem and introduce the Causal
Improvement Graph (CIG), which maintains and updates a persistent improvement state. Figure~\ref{fig:cig-overview} summarizes the state
transition and the information represented by each typed node, which we detail through the
state representation, discovery operations, and outcome-grounded update.

\vspace{-6pt}
\subsection{Harness Optimization}

A Harness is the executable runtime that constructs task context and controls an agent's
execution flow. We study the problem of discovering an effective Harness under a fixed model
and external evaluation. A proposer is an LLM-based agent that uses prior candidate artifacts
and evaluation feedback to propose and implement Harness changes. The discovery process
alternates candidate construction with external evaluation of the resulting Harnesses. Let
$\mathcal{H}$ denote the space of possible Harnesses and let $\mathcal{H}_0 \subset \mathcal{H}$
be an initial set. Let $B$ denote the candidate-evaluation budget, measured in complete candidate
evaluations. A discovery method constructs a candidate set $\mathcal{C}_B \subset \mathcal{H}$
within this budget. The evaluation function $\mathcal{E}$ maps each
candidate Harness $h \in \mathcal{H}$ to its measured behavior on a task distribution $\mathcal{D}$,
\begin{equation}
    \mathbf{y}(h) = \mathcal{E}(h,\mathcal{D}) \in \mathcal{Y}.
    \label{eq:evaluation}
\end{equation}
The outcome space $\mathcal{Y}$ may contain a single task score or multiple dimensions such as
quality, reliability, cost, and latency. A selection policy $\Pi$, fixed before discovery,
chooses the final Harness from the evaluated candidates,
\begin{equation}
    \hat{h}_B = \Pi\!\left(\left\{(h,\mathbf{y}(h)) \mid h\in\mathcal{C}_B\right\}\right).
    \label{eq:harness-selection}
\end{equation}
This formulation accommodates different selection policies without defining Harness quality
as a single scalar. While $\Pi$ selects the final Harness, discovery must decide which
changes to try within the evaluation budget. This requires using earlier experiments to
judge which directions remain promising and which judgments need revision.

\newpage
\subsection{Causal Improvement State}

Proposer-centric methods use accumulated candidate artifacts, evaluations, and diagnostic
records to propose further changes. These records preserve accumulated experience, but the proposer
must reconstruct how prior findings relate and whether earlier judgments still hold when
planning the next change. CIG makes these relationships and current judgments explicit in a
persistent improvement state. Linking observations recorded as evidence to hypotheses, interventions, and outcomes
allows a new result to revise the judgments it bears on, while preserving their experimental
basis. Subsequent proposals can then build on the updated state, connecting what has been
observed to what to improve next and which earlier judgments need revision.

We represent this state at discovery round $t$ as a committed causal improvement graph
$G_t=(V_t,R_t)$. We use \emph{causal} in the operational sense of intervention-grounded
improvement reasoning. Formal causal identification of the task distribution is outside this
formulation. The four semantic node types of
$G_t$ are \evidenceterm, \hypothesisterm, \interventionterm, and
\outcometerm. Harness artifacts and plans are associated records linked to these nodes
through references. Its relations $R_t$ record provenance, support, contradiction, and Harness lineage.
A basic causal path in the graph is
\begin{equation}
    \text{\evidencetag} \longrightarrow \text{\hypothesistag}
    \longrightarrow \text{\interventiontag} \longrightarrow \text{\outcometag}.
    \label{eq:logic-path}
\end{equation}
Concretely, an \evidenceterm node stores an observed behavior with its task and Harness
provenance. A linked \hypothesisterm node states a falsifiable explanation. An
\interventionterm node specifies the executable Harness change intended to test it, and an
\outcometerm node records the measured behavior and its verdict. These measurements come
from external evaluation, which executes the candidate
Harness and measures its behavior using $\mathcal{E}$. These results supply evidence for the
next discovery round. Across successful and unsuccessful experiments, $G_t$ retains the
observations, hypotheses, interventions, and outcomes together with their logical relations,
source references, and current hypothesis assessments. This structured record forms the
persistent improvement state. Subsequent operations can trace a proposed change to its
supporting observations and consult updated assessments of related hypotheses, making accumulated
experience directly usable in planning further Harness optimization from linked observations
and current hypothesis assessments.

\subsection{Graph-Governed Discovery Loop}

CIG advances the persistent improvement state represented by $G_t$ through graph-governed
discovery operations. Each discovery round updates $G_t$ to $G_{t+1}$ through the following sequence:
\begin{equation}
    G_t \xrightarrow{\textsc{Observe}} A_t
    \xrightarrow{\textsc{Plan}} P_t
    \xrightarrow{\textsc{PlanSelect}} S_t
    \xrightarrow{\textsc{Implement}} \widetilde{\mathcal H}_t
    \xrightarrow{\textsc{Evaluate}} O_t
    \xrightarrow{\textsc{Update}} G_{t+1}.
    \label{eq:graph-loop}
\end{equation}
The intermediate variables denote linked working products within this transition:
an analysis $A_t$ of the current graph state, a plan set $P_t$, a selected portfolio
$S_t$, realized Harnesses $\widetilde{\mathcal H}_t$, and grounded \outcometerm nodes $O_t$.
\textsc{Observe} reads the current graph and permitted raw history to summarize what has been
learned and what remains unclear.
Each output cites addressable evidence, so a later plan can recover the observations behind it.
\textsc{Plan} proposes falsifiable \hypothesisterm--\interventionterm pairs with named parent
Harnesses, preserved mechanisms, and expected outcomes.
\textsc{PlanSelect} uses a separate proposer call to assess every plan on seven 0--4 scales:
evidence strength, expected gain, information value, novelty, falsifiability, cost, and risk.
It selects a portfolio using these assessments and pairwise redundancy, including both an
evidence-grounded refinement and a branch or restart when the selected budget permits.
The selector records its selection rationale and rejection reasons.
In Full CIG, these judgments use the current relations to connect each proposed intervention to
its evidence and hypothesis, including unresolved or revised conclusions.
Independent
\textsc{Implement} operations realize the selected plans as executable Harnesses.
\textsc{Evaluate} measures their behavior, and \textsc{Update} incorporates the grounded
\outcometerm nodes into the graph and revises hypothesis assessments.
Through these operations, CIG connects each evaluated change to the state used for subsequent planning, allowing
later proposals to build on prior results and revised judgments across discovery rounds.

Proposers are invoked for locally scoped operations. They interpret \evidenceterm nodes, formulate
\hypothesisterm nodes, or implement a selected \interventionterm node, but they do not own the persistent state or the
complete discovery process. The graph supplies each invocation with a role, a local objective,
and explicit dependencies on prior work. Algorithm~\ref{alg:cig} summarizes the resulting loop.

\newpage
\subsection{Outcome-Grounded Update}
\begin{wrapfigure}{r}{\dimexpr0.55\textwidth-\baselineskip\relax}
    \vspace{-24pt}
    \hrule
    \vspace{3pt}
    \refstepcounter{algorithm}
    \noindent\textbf{Algorithm~\thealgorithm: CIG discovery loop}
    \label{alg:cig}
    \vspace{3pt}
    \hrule
    \vspace{2pt}
    \normalsize
    \algrenewcommand\algorithmicindent{0.8em}
    \begin{algorithmic}[1]
        \Require Initial Harnesses $\mathcal{H}_0$, evaluation $\mathcal{E}$, policy $\Pi$, budget $B$
        \Ensure Selected Harness $\hat h_B$, final graph $G_T$
        \State $G_0 \gets \Call{Initialize}{\mathcal{H}_0}$
        \For{$t=0,1,\ldots$ while budget remains}
            \State $A_t \gets \Call{Observe}{G_t}$
            \State $P_t \gets \Call{Plan}{G_t,A_t}$
            \State $S_t \gets \Call{PlanSelect}{G_t,P_t,B}$
            \ForAll{$p \in S_t$ \textbf{in parallel}}
                \State $h_{t,p} \gets \Call{Implement}{p,G_t}$
                \State $\mathbf{y}_{t,p} \gets \Call{Evaluate}{h_{t,p},\mathcal{E}}$
                \State $\tau_{t,p} \gets \Call{Compare}{h_{t,p},p,\mathbf{y}_{t,p}}$
            \EndFor
            \State $G_{t+1} \gets \Call{Update}{G_t,P_t,S_t,\{\tau_{t,p}\}}$
        \EndFor
        \State $\hat h_B \gets \Call{HarnessSelect}{G_T,\Pi}$
        \State \Return $\hat h_B,G_T$
    \end{algorithmic}
    \hrule
\end{wrapfigure}

The final \textsc{Update} step in each discovery round (Equation~\ref{eq:graph-loop}) incorporates
externally evaluated candidate behavior into the persistent improvement state through a structured
graph transition. For candidate $h_{t,p}$ and its parent Harness, CIG records
\begin{equation}
    \tau_{t,p} =
    \left(\mathbf{y}_{t,p},\mathcal{T}^{+}_{t,p},\mathcal{T}^{-}_{t,p},
    \mathcal{T}^{0}_{t,p},e_{t,p},a_{t,p},v_{t,p}\right).
    \label{eq:transition-ledger}
\end{equation}

The sets $\mathcal{T}^{+}_{t,p}$, $\mathcal{T}^{-}_{t,p}$, and $\mathcal{T}^{0}_{t,p}$ record
corrected, regressed, and unchanged outcomes relative to the parent.
Intervention fidelity $e_{t,p}$ records whether the intended change was executed,
association $a_{t,p}$ compares its effects with the predicted direction, and verdict $v_{t,p}$
assesses the linked hypothesis at the scope supported by the observations.
Unavailable execution or task-transition evidence remains unspecified.
Appendix~\ref{app:verdict-semantics} distinguishes score-based judgments from evidence about
the underlying mechanism.
The update operator appends these transitions and their dependencies to the persistent state,
\begin{equation}
    G_{t+1}=\operatorname{Update}\!\left(G_t,
    \{\tau_{t,p}\mid p\in S_t\}\right).
    \label{eq:graph-update}
\end{equation}
An \outcometerm node serves two roles: measured performance informs Harness selection,
while the linked prediction and observed effects update discovery knowledge.
Revision separates measured results from their interpretation. The evaluator supplies scores
and parent-relative changes, while reasoning over linked evidence and hypotheses compares these observations with
the linked prediction and revises its current assessment. Conflicting results can narrow the
supported scope or leave an explanation unresolved. The graph retains the original experiment
and its provenance alongside the assessment used for subsequent observation and planning.

Subsequent observation and planning read these assessments together with their evidence.
An unsuccessful candidate can therefore expose an informative trade-off or challenge an earlier
explanation. Performance selection follows the fixed eligibility policy, independently of
whether every proposed explanation is supported. Appendix~\ref{app:operation-details} describes
the retained records and current hypothesis assessments.

\par\medskip
\noindent\textbf{In summary,} CIG externalizes the evolving improvement state in a persistent
graph. Explicit relations connect evidence, hypotheses, interventions, and outcomes, while
externally evaluated outcomes revise hypothesis assessments across iterations. This allows
local proposers to build directly on prior experimental dependencies and updated assessments
rather than recover them from raw history, enabling prior findings to directly govern
subsequent Harness optimization.

\section{Experiments}
\label{sec:experiments}

In this section, we investigate five questions to assess CIG's effectiveness across environments
and models and understand how its search allocation, structural components, and outcome-grounded
revisions contribute to automated Harness optimization.

\begin{tcolorbox}[
  enhanced, colback=cyan!4, colframe=cyan!45!blue!60,
  boxrule=0pt, sharp corners,
  borderline west={3pt}{0pt}{cyan!55!blue!65},
  left=10pt, right=9pt, top=7pt, bottom=7pt,
  before skip=7pt, after skip=8pt]
\small\itshape
\begin{description}[leftmargin=3.2em,labelwidth=2.8em,labelsep=0.4em,
  font=\color{cyan!55!blue!65}\bfseries\itshape,
  itemsep=4pt,topsep=0pt,parsep=0pt]
\item[RQ1:] How effective is CIG compared with existing approaches?
\item[RQ2:] How do CIG's search schedules perform across solvers?
\item[RQ3:] Does CIG remain effective across proposer models?
\item[RQ4:] Which components contribute to CIG's improvements?
\item[RQ5:] How does CIG turn outcomes into revisable knowledge?
\end{description}
\end{tcolorbox}

\subsection{Evaluation Protocol}

\noindent\textbf{Benchmarks.}
Our classification studies use three tasks from Meta-Harness~\citep{lee2026metaharness}:
USPTO-50k reaction classification~\citep{schneider2016reaction}, diagnosis from symptom descriptions using Gretel's adaptation of
Symptom2Disease~\citep{gretel2023symptom}, and LawBench criminal-charge
classification~\citep{fei2024lawbench}. Our agent-Harness evaluation covers SWE-bench Verified
repository repair~\citep{jimenez2024swebench}, stateful application interaction in
AppWorld~\citep{trivedi2024appworld}, and terminal-based tasks in Terminal-Bench~\citep{merrill2026terminalbench}.
Table~\ref{tab:main-benchmark-results} compares CIG with published results under the HarnessFix
evaluation protocol. Our runs reuse the benchmark environments and evaluation setup, with
aligned initial Harnesses, split sizes, task and optimization models, and three-run reporting.
Sampling, data splits, search schedules, and budgets follow the HarnessFix protocol.

\noindent\textbf{Protocol.}
Within each controlled comparison, methods share initialization, access to raw observations and
evaluation history, and the evaluation allowance. The agent-benchmark comparison uses the
HarnessFix protocol described above, while the subsequent matched-control studies use eight
candidate evaluations. Candidate selection depends only on validation performance.
Held-out test outcomes are revealed only after search and selection are complete. Our primary
measure is the held-out test performance of the validation-selected Harness. Selection regret is the difference between the test oracle and the test score of the
validation-selected Harness. Search curves summarize progress during discovery.

\noindent\textbf{Comparison methods.}
We compare CIG with (1) prompt and scaffold optimization methods and (2) Harness code
optimization methods. In the first group, GEPA~\citep{agrawal2026gepa} and
SCOPE~\citep{pei2025scope} evolve prompts through reflective or memory-guided feedback,
while ReCreate~\citep{hao2026recreate} converts experience into reusable agent scaffolds.
In the second group, Meta-Harness~\citep{lee2026metaharness} searches complete Harness code
from prior artifacts, whereas HarnessFix~\citep{chen2026harnessfix} diagnoses failed
trajectories and applies scoped repairs.
To evaluate CIG as an integrated discovery method, we additionally implement
\textbf{Proposer-Centric Full-History Search (Full History)} as a graph-free control. It gives one
proposer the complete permitted history and a skill governing history inspection and candidate implementation while
matching CIG in initialization, observable evidence, and candidate-evaluation budget. This
comparison tests the complete CIG workflow against proposer-centric search.

\Needspace{7\baselineskip}
\subsection{\rqmark{1}: How effective is CIG compared with existing approaches?}
\label{sec:rq1}

\begin{table*}[t]
    \caption{Held-out test performance of optimized Harnesses under the HarnessFix evaluation protocol. Task and optimization
    agents use the GPT-5 mini configuration~\citep{openai2025gpt5mini}. Entries are test task-completion rates (\%), averaged over three runs. Published rows reproduce HarnessFix's reported
    results, and our runs use the aligned evaluation setup. Full History is our baseline,
    matched to CIG in initialization and evaluation allowance. Parentheses in the CIG row
    show percentage-point gains over the reproduced initial Harness.
    Reporting conventions are given in Appendix~\ref{app:configuration}.}
    \label{tab:main-benchmark-results}
    \centering
    \small
    \setlength{\tabcolsep}{3.5pt}

    \begin{tabular}{l*{4}{c@{\hspace{1pt}}l}}
        \toprule
        Method & \makebox[0pt]{SWE-V} & & \makebox[0pt]{AppWorld} &
        & \makebox[0pt]{TB2.0-V} & & \makebox[0pt]{Macro avg.} & \\
        \midrule
        \TableOneRows
        \bottomrule
    \end{tabular}
\end{table*}

\begin{table}[!t]
    \centering
    \caption{Robustness across task solvers on the three classification tasks.
    Entries are mean test scores (\%) and selection regret under eight candidate evaluations.
    $\Delta$ reports test gains over Full History in percentage points.
    Wide uses two waves of four candidates, and Deep uses four waves of two.}
    \label{tab:solver-robustness}
    \centering
    \resizebox{0.94\textwidth}{!}{%
    \begin{minipage}{1.04\textwidth}
    \footnotesize
    \setlength{\tabcolsep}{0pt}
    \renewcommand{\arraystretch}{1.15}
    \begin{tabular}{@{}p{0.23\linewidth}>{\centering\arraybackslash}p{0.09\linewidth}@{\hspace{0.01\linewidth}}*{2}{>{\centering\arraybackslash}p{0.09\linewidth}}@{\hspace{0.015\linewidth}}*{2}{>{\centering\arraybackslash}p{0.09\linewidth}}@{\hspace{0.015\linewidth}}>{\centering\arraybackslash}p{0.09\linewidth}@{\hspace{0.01\linewidth}}*{2}{>{\centering\arraybackslash}p{0.09\linewidth}}@{}}
        \toprule
        & \multicolumn{3}{c}{Val-selected Test $\uparrow$}
        & \multicolumn{2}{c}{$\Delta$ Test $\uparrow$}
        & \multicolumn{3}{c}{Selection regret $\downarrow$} \\
        \cmidrule(lr){2-4}\cmidrule(lr){5-6}\cmidrule(lr){7-9}
    Solver & \makebox[0pt]{Full History} & Wide & Deep & Wide & Deep & \makebox[0pt]{Full History} & Wide & Deep \\
        \midrule
        \TableTwoRows
        \bottomrule
    \end{tabular}
    \end{minipage}
    }
\end{table}

\begingroup
\setlength{\intextsep}{6pt}
\begin{wrapfigure}{R}{0.60\textwidth}
    \setlength{\abovecaptionskip}{2pt}
    \setlength{\belowcaptionskip}{0pt}
    \raggedright
    \includegraphics[width=\linewidth]{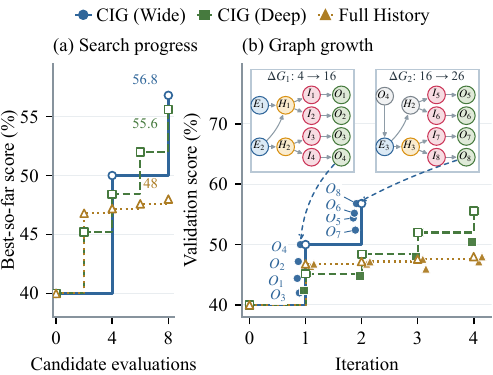}
    \caption{Search progress and graph growth on LawBench using post-search validation re-evaluation means. (a) Best-so-far scores at completed wave boundaries. (b) Filled markers show individual candidate scores, linked to Outcome nodes; hollow markers and step curves show the best-so-far score, as in (a). Insets show node counts before and after updates, with gray marking retained nodes and colors marking additions. Iteration zero denotes the initial graph state before the first round of Harness optimization.}
    \label{fig:search-dynamics}
    \vspace{-8pt}

\end{wrapfigure}

Table~\ref{tab:main-benchmark-results} reports higher CIG scores than both our Full History
control and the published baselines across repository repair, application interaction, and
terminal tasks. The gains over Full History across these environments support the complete
method beyond a single task interface. The following controlled studies examine model
dependence, search allocation, and component contributions.

\Needspace{7\baselineskip}
\subsection{\rqmark{2}: How do CIG's search schedules perform across solvers?}
\label{sec:rq2}

We fix GPT-5.4 as the proposer for all three classification tasks:
USPTO-50k, Symptom2Disease, and LawBench. To assess whether the gains depend on the
task model, we vary the solver used to execute each Harness.
We use five solvers, including two models from the Qwen family:
Qwen3.5-Plus~\citep{alibabagroup2026qwen35} and Qwen3.7-Flash~\citep{qwenteam2026qwen37flash}.
The other three are DeepSeek-V4-Flash~\citep{deepseekai2026v4},
GLM-5.2-Fast-Preview~\citep{alibabacloud2026glm52fast}, and MiniMax-M2.5~\citep{minimax2026m25}.
We compare Full History with two CIG search settings. CIG (Wide)
uses two waves of four validated candidates, while CIG (Deep) uses four waves of two validated
candidates. Full History shares Deep's four-wave, two-candidate schedule.
Table~\ref{tab:solver-robustness} first compares Deep with Full History under this matched
schedule. Deep improves on Full History for four of the five solvers, with a small decrease
for GLM, supporting the integrated CIG workflow beyond search width.
Within CIG, Wide outperforms Deep for every solver and has lower mean selection regret.
This favors broader candidate portfolios with fewer feedback rounds in the tested settings.
RQ4 examines component contributions under
matched scheduling and staged calls.

\begin{wraptable}{r}{0.57\textwidth}
    \vspace{-10pt}

    \caption{Robustness across proposers on SWE-bench Verified. Scores are mean test pass@1
    $\pm$ sample standard deviation of the validation-selected Harness.
    $\Delta$ is the difference in means.}
    \label{tab:proposer-dependence}
    \centering
    \small
    \setlength{\tabcolsep}{3pt}
    \renewcommand{\arraystretch}{1.05}
    \begin{tabular*}{0.98\linewidth}{@{\extracolsep{\fill}}l*{3}{>{\small}c}@{}}
        \toprule
        Proposer & \small Full History & \small CIG (Wide) & \small $\Delta$ \\
        \midrule
        \TableThreeRows
        \bottomrule
    \end{tabular*}
    \vspace{-6pt}
\end{wraptable}

Figure~\ref{fig:search-dynamics} compares candidates discovered under the same
eight-candidate allowance using post-search re-evaluation means. Full History shows modest
later gains, whereas both CIG schedules discover stronger candidates as search progresses.
Deep exceeds Full History's final re-evaluated best score after two waves and improves further,
while Wide reaches the highest final score. The insets show how later experiments build on
shared evidence and retained hypotheses within this evaluated search horizon,
connecting results across discovery rounds.
\par\WFclear
\endgroup
\newpage
\begin{wrapfigure}{r}{0.55\textwidth}
    \vspace{6pt}
    \setlength{\abovecaptionskip}{2pt}
    \centering
    \includegraphics[width=\linewidth]{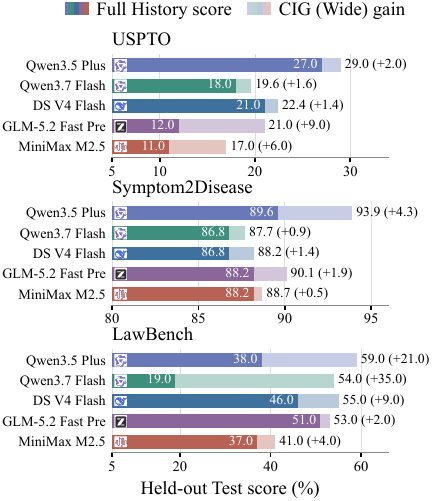}
    \caption{Mean task-level test performance. Labels show CIG (Wide) scores and gains over Full History in percentage points. USPTO denotes USPTO-50k, DS denotes DeepSeek, and Pre denotes Preview. Each panel uses its own linear axis with a nonzero origin. Bar lengths are comparable within each panel, but not across panels with different axis ranges.}
    \label{fig:solver-task-gains}
    \vspace{-46pt}
\end{wrapfigure}

Figure~\ref{fig:solver-task-gains} shows that the aggregate advantage extends across the three
classification tasks: gains averaged across solvers are positive on each task, and the
task-level comparisons are positive across all five solvers. Their magnitudes vary, but the
consistent direction supports robustness to solver choice across all three classification tasks
and five evaluated solver models.

\noindent\begin{minipage}{\dimexpr0.45\textwidth-\columnsep\relax}
\subsection{\rqmark{3}: Does CIG remain effective across proposer models?}
\end{minipage}\par
\label{sec:rq3}

We vary the proposer on SWE-bench Verified, where Harness design coordinates repository
navigation, context construction, tool use, and repair strategy. We fix Qwen3.5-Plus~\citep{alibabagroup2026qwen35} as the task solver and compare Full History with
CIG (Wide) using GPT-5.5~\citep{openai2026gpt55}, GPT-5.6 Sol~\citep{openai2026gpt56},
GLM-5.3~\citep{zai2026glm53}, Kimi K3~\citep{kimiteam2026kimik3}, and
DeepSeek-V4-Pro~\citep{deepseekai2026v4} as proposers. The Harness initialization, task split, selection rule, and
eight-candidate evaluation budget remain fixed. Table~\ref{tab:proposer-dependence} shows a
positive gain with every proposer, indicating that the benefit is not confined to a single
proposal model. The variation in gain motivates the matched structural comparisons under a common
discovery workflow.

\Needspace{7\baselineskip}
\subsection{\rqmark{4}: Which components contribute to CIG's improvements?}
\label{sec:rq4}

We next examine which parts of CIG account for its gains. Table~\ref{tab:structural-ablation}
combines cumulative construction with targeted alternatives to distinguish the benefits of
organizing experimental history, linking records, revising assessments, and using that state
to select plans. Holding the discovery workflow fixed allows these contrasts to assess how
the components contribute to final Harness quality and search progress under the same evaluation allowance.

\noindent\textbf{State and relation semantics.}
M0 exposes raw history through staged calls. M1 organizes it into typed E--H--I--O records,
and M2 adds their logical relations. Both improve test and AUC for both proposers.
B1 retains M2's structure but shuffles its relations and loses performance, supporting the
value of meaningful dependencies beyond typed records alone.

\noindent\textbf{Persistence and revision.}
M3 replaces M2's round-local state with persistent, revisable assessments. B2 preserves
persistence but permits only append-only records. M3 outperforms both M2 and B2,
indicating that carrying knowledge forward and updating its interpretation make complementary
contributions along this construction path.

\noindent\textbf{Using relational state for selection.}
M4 adds graph-guided plan selection to M3 and achieves the highest test and AUC with both
proposers. B3 retains persistent, revisable records and local provenance but removes explicit
cross-record relations, so selection uses the records directly. Its performance lies between
M0 and M4: revisable notes retain substantial value, while cross-record relations provide a
further benefit. Together, these comparisons suggest a practical design principle: maintain
both accumulated experience and its evolving interpretation as inputs to Harness optimization.
Revisable assessments track what prior results currently support, while explicit relations
connect those judgments to subsequent planning and plan selection.
Appendices~\ref{app:stage-three} and~\ref{app:stage-three-variance} provide full control definitions
and run-to-run standard deviations.

\begin{table}[!t]
    \caption{Structural contributions to CIG. Starting from Full History, cumulative variants
    introduce typed state, relations, and revision before full CIG, while alternative designs
    (blue) test how these components are used. The results support their complementary value
    for final Harness quality and search progress under a matched workflow.
    Appendix~\ref{app:stage-three} provides exact variant definitions and reporting conventions.}
    \label{tab:structural-ablation}
    \centering
    \small
    \setlength{\tabcolsep}{2.5pt}
    \renewcommand{\arraystretch}{1.15}
    \resizebox{\textwidth}{!}{%
    \begin{tabular}{@{}lccccc@{\hspace{3pt}}l@{\hspace{3pt}}c@{\hspace{10pt}}c@{\hspace{3pt}}l@{\hspace{3pt}}c@{\hspace{4pt}}}
        \toprule
        & \multicolumn{4}{c}{Cumulative construction}
        & \multicolumn{3}{c}{Proposer A}
        & \multicolumn{3}{c}{Proposer B} \\
        \cmidrule(lr){2-5}\cmidrule(lr){6-8}\cmidrule(lr){9-11}
        Variant & State & Relations & Update & Selection
        & \makebox[0pt]{\hspace{6pt}Test $\uparrow$} & & \makebox[0pt]{AUC $\uparrow$}
        & \makebox[0pt]{\hspace{6pt}Test $\uparrow$} & & \makebox[0pt]{AUC $\uparrow$} \\
        \midrule
        \TableFourRows
        \bottomrule
    \end{tabular}%
    }
\end{table}

\Needspace{4\baselineskip}
\vspace{-6pt}
\subsection{\rqmark{5}: How does CIG turn outcomes into revisable knowledge?}
\label{sec:rq5}

The following trace shows how CIG connects a diagnosed weakness to a testable intervention and
a revised assessment. The node excerpts preserve the reasoning chain across two discovery waves.

\begin{tcolorbox}[
    enhanced, unbreakable, colback=white, colframe=black!35,
    colbacktitle=black!5, coltitle=black,
    title={Evidence-linked intervention and reassessment}, fonttitle=\bfseries,
    boxrule=0.55pt, arc=2pt, left=8pt, right=8pt, top=3pt, bottom=3pt,
    before skip=3pt, after skip=3pt]
\small
\setlength{\smallskipamount}{1pt}
\centering
$\textcolor{evidencebar}{E_2}
\xrightarrow{\text{grounds}}\textcolor{hypothesisbar}{H_2}
\xrightarrow{\text{tested by}}\textcolor{interventionbar}{I_2}
\xrightarrow{\text{yields}}\textcolor{outcomebar}{O_2}$
\par\smallskip
\raggedright
\evidencetag\ $E_2$ \textbf{Identify a trade-off.}\par
Similarity retrieval improves USPTO and Symptom2Disease but lowers LawBench from $60$ to $46$.
\par\smallskip
\hypothesistag\ $H_2$ \textbf{Register a testable explanation.}\par
Restoring demonstration coverage should recover LawBench while preserving the other gains
and exceeding the parent's aggregate score of $58.00$.
\par\smallskip
\interventiontag\ $I_2$ \textbf{Test a scoped change.}\par
Keep query-conditioned retrieval, allocate 60\% of the prompt budget to nearest neighbors,
and backfill the remaining slots using answer-space and token-coverage novelty.
\par\smallskip
\outcometag\ $O_2$ \textbf{Separate effects from the joint verdict.}\par
LawBench recovers to $52$ and Symptom2Disease stays at $88$, but USPTO falls to $33.33$.
Aggregate performance falls to $57.78$. The joint target fails, while the predicted LawBench
direction remains consistent with the observation.
\par\smallskip
\centering
$\textcolor{outcomebar}{O_2}\xrightarrow{\text{updates assessment}}\textcolor{hypothesisbar}{H_2}$\par
\textit{Local prediction remains consistent. Joint target fails. Coverage explanation remains open.}
\end{tcolorbox}

The linked state preserves both the local recovery and the failed joint target, leaving a
scoped assessment that guides subsequent planning.
Appendix~\ref{app:case-trace} provides the detailed trace and a state-intervention comparison
that contrasts the plans and Harnesses produced when hypothesis assessments are updated using
evaluated outcomes or retained unchanged.

\vspace{-8pt}
\section{Conclusion}

CIG shifts Harness optimization from proposer-centric to graph-governed discovery through
persistent improvement state, allowing proposers to reuse accumulated experience and
updated assessments rather than reconstruct them from raw history. Experiments show stronger
Harnesses across tasks and robustness to solver and proposer choice. Matched ablations support
explicit relations beyond revisable records, while a state-intervention study shows how updated
assessments guide subsequent plans. Together, these findings support maintaining accumulated
experience and its evolving interpretation as persistent improvement state to guide further
Harness optimization. Our evaluation focuses on Harness optimization within individual searches,
leaving the transfer and adaptation of persistent improvement state across tasks and models as
a promising direction for future work.

\appendix
\clearpage
\section*{Appendix}
\startcontents[sections]
\section*{Table of Contents}
\begingroup
\hypersetup{linkcolor=darkblue}
\printcontents[sections]{l}{1}{\setcounter{tocdepth}{2}}
\endgroup
\hypersetup{linkcolor=mylinkcolor}
\bigskip

\section{Additional Method and Experimental Details}
\label{app:details}

\subsection{Graph Schema and Transition Rules}

The graph contains four independently addressable node types: Evidence, Hypothesis, Intervention,
and Outcome. Evidence retains the source observation and its task and Harness provenance.
A Hypothesis has its own identity, explanatory claim, applicable scope, and revisable assessment.
Plans reference these nodes and specify how an Intervention will test the linked Hypothesis.
Outcomes retain the external measurements and connect them to the executed Intervention.
Harness references record the parent and resulting candidate without replacing the four node types.

Relations express which Evidence supports or challenges a Hypothesis, which Intervention tests
it, which Outcome evaluates that Intervention, and which earlier assessment the result revises.
Multiple plans can test the same Hypothesis, and one Outcome can inform several linked assessments.
When an Outcome supplies Evidence for a later round, subsequent reasoning references the same
measurement rather than counting it as a new independent observation.

Observe produces evidence-linked claims, uncertainty, unresolved questions, and cross-branch
conflicts. Plan proposes interventions with parents, preserved mechanisms, predicted effects,
falsification conditions, and resource contracts. PlanSelect scores each plan on seven ordinal
criteria and compares portfolio redundancy and complementarity. Appendix~\ref{app:operation-details}
specifies the selection interface. Implement realizes the selected changes, while external evaluation
produces the measured outcomes. Update appends provenance and deltas and refreshes the decision
view for subsequent calls. Related hypotheses can remain distinct, with shared evidence made
explicit through their relations. Search ends when the candidate-evaluation allowance is exhausted,
which bounds graph growth in these experiments.

\subsection{Related Methods and Representation Choices}
\label{app:related-comparison}

Table~\ref{tab:related-features} compares how related methods retain experience, represent
predictions, and organize optimization state.

\begin{table}[!t]
\centering
\caption{State representations in related methods. Columns indicate persistent search or
experience memory, explicit predictions or hypotheses, and graph-structured state used in search
or adaptation. The final column contrasts each method's organizing principle with CIG.
\featureyes\ denotes an explicitly described feature, and \featureunspecified\ denotes a feature
not established by the cited description.}
\label{tab:related-features}
\footnotesize
\setlength{\tabcolsep}{3pt}
\renewcommand{\arraystretch}{1.12}
\begin{tabularx}{\textwidth}{@{}>{\raggedright\arraybackslash}p{0.25\textwidth}cccY@{}}
\toprule
\textbf{Method} & \shortstack{\textbf{Persistent}\\\textbf{memory}} & \shortstack{\textbf{Explicit}\\\textbf{predictions}} & \shortstack{\textbf{Graph}\\\textbf{state}} & \textbf{Compared with CIG} \\
\midrule
ADAS~\citep{hu2024adas} & \featureyes & \featureunspecified & \featureunspecified & Organizes search around archived programs, while CIG organizes it around revisable hypotheses. \\
AFlow~\citep{zhang2024aflow} & \featureyes & \featureunspecified & \featureyes & Searches a workflow tree, while CIG links evidence and hypotheses across interventions. \\
Meta-Harness~\citep{lee2026metaharness} & \featureyes & \featureunspecified & \featureunspecified & A proposer interprets stored history, while CIG exposes linked assessments to scoped operations. \\
AHE~\citep{lin2026ahe} & \featureyes & \featureyes & \featureunspecified & Checks edit-level predictions, while CIG carries their assessments into a persistent hypothesis graph. \\
HarnessFix~\citep{chen2026harnessfix} & \featureyes & \featureyes & \featureunspecified & Organizes repair memory, while CIG uses linked hypothesis assessments to select experiments. \\
Living-Harness~\citep{du2026livingharness} & \featureyes & \featureunspecified & \featureyes & Updates procedural behavior, while CIG updates the optimizer's experimental reasoning state. \\
HSI~\citep{zhou2026hsi} & \featureyes & \featureyes & \featureyes & Evolves the Harness and evolver, while CIG revises hypothesis state through fixed typed operations. \\
AutoSaddler~\citep{park2026autosaddler} & \featureyes & \featureunspecified & \featureyes & Centers its graph on patch lineage and lessons, while CIG centers it on revisable hypotheses. \\
EviGraph (research)~\citep{ren2026evigraph} & \featureyes & \featureyes & \featureyes & Links evidence to research claims, while CIG links it to evaluated Harness interventions. \\
EviGraph (information seeking)~\citep{chen2026evigraph} & \featureyes & \featureunspecified & \featureyes & Grounds information seeking in evidence, while CIG uses evidence to guide Harness experiments. \\
\midrule
\textbf{CIG} & \featureyes & \featureyes & \featureyes & Linked, outcome-revised hypotheses govern subsequent Harness experiments. \\
\bottomrule
\end{tabularx}
\end{table}

\paragraph{Representation and update targets.}
ADAS and AFlow retain program archives and search trees, while AHE and HarnessFix organize
predictions and repair experience. Among graph-based systems, AutoSaddler records patch lineage
and lessons, HSI also evolves the optimizer, and Living-Harness updates procedural behavior.
The two EviGraph systems organize research claims and information-seeking evidence, respectively.
\textbf{CIG} makes revisable hypotheses the interface between evidence, executable Harness
interventions, and subsequent plan selection. Its distinction is the use of these linked
assessments to govern discovery, rather than graph storage alone.
Table~\ref{tab:structural-ablation} isolates representation, revision, and selection mechanisms.

\paragraph{Evaluation methodology.}
Rethinking Harness Evolution compares automatic evolution with simple search under matched
feedback and inference budgets and evaluates held-out generalization~\citep{wang2026rethinkingharness}.
Table~\ref{tab:main-benchmark-results} provides a comparison under the HarnessFix protocol.
Our subsequent matched controls address method behavior under a fixed candidate-evaluation
allowance, and Table~\ref{tab:structural-ablation} also matches the staged reasoning schedule.

\subsection{Structural Controls and Interpretable Contrasts}
\label{app:stage-three}

Table~\ref{tab:structural-ablation} reports a cumulative construction and diagnostic
branches. The definitions below specify what changes in each comparison.

\paragraph{Reporting conventions.}
Proposers A and B denote GPT-5.6 Sol and DeepSeek-V4-Pro, respectively, with Qwen3.7-Flash~\citep{qwenteam2026qwen37flash}
as solver on SWE-bench Verified. Values are five-run means. Test is validation-selected
held-out pass@1 (\%), and AUC summarizes the best-so-far validation curve. Parentheses
report test differences in percentage points: M1--M4 versus the preceding main variant,
and B1, B2, and B3 versus M2, M3, and M4, respectively.

\paragraph{Matched protocol.}
M0 is a staged raw-history control for structural attribution. It uses the same staged calls, round
schedule, portfolio size, and final Harness selection rule as M1--M4. Across the complete matrix,
we also fix the initialization, solver, benchmarks, candidate-evaluation budget, and access to
eligible raw history. The variants differ only in the state interface, relation semantics,
cross-round update, or selection access named in the table. Candidate generation and selection
use validation only, and held-out test remains
unavailable until the selections are frozen.

\paragraph{M0 Full History.} Supply the staged operations with the visible raw discovery history. No
compiled E--H--I--O state or logic relation is carried between rounds.
\paragraph{M1 Typed State.} Compile the same eligible observations into E--H--I--O records without
explicit logic relations. The compiled state is reset before the next round, so this step tests
the typed representation rather than persistence.
\paragraph{M2 Logic Relations.} Add typed relations that connect Evidence, Hypotheses, Interventions,
and Outcomes within the current round. The graph is still reset before the next round, and plan
selection does not directly traverse the relations.
\paragraph{B1 Shuffled Relations.} Replace the correct relations of M2 with a recorded type-compatible
shuffle. Node content, source attribution, evaluation identity, and executable lineage remain
unchanged. The comparison with M2 tests relation semantics rather than edge count.
\paragraph{M3 Persistent Revision.} Carry the relational state across rounds. New Outcomes can revise
the status or scope of earlier Hypotheses and can redirect subsequent Interventions. Selection
remains independent of direct graph access.
\paragraph{B2 Append-only Persistence.} Carry the M2 graph across rounds but only append records.
Earlier conclusions are not revised when later Outcomes conflict with them. The comparison with
M3 separates durable storage from revision-aware experimental state.
\paragraph{M4 Full CIG.} Give plan selection access to the current revision-aware graph so the selected
portfolio can target unresolved hypotheses, competing explanations, and supported directions.

\paragraph{B3 Without Cross-record Relations.}
This control starts from M4 and preserves typed E--H--I--O records, cross-round persistence,
hypothesis status and scope revision, and access to the resulting records during selection.
It retains the same staged calls, plan and candidate counts, initialization, evaluation protocol,
and raw-history access. Each experiment retains its evidence sources, tested hypothesis,
intervention, candidate, and outcome references, together with revision audit records.
The control does not construct, persist, or expose cross-record semantic relations such as
support and contradiction across experiments. These relations are not copied into summaries
or alternative adjacency fields. Observe, Plan, and PlanSelect instead read the persistent records
and their current assessments. The proposer can infer connections from their contents, and
Update can revise earlier assessments from new outcomes without storing a semantic edge network.
Thus B3 removes explicit relational organization, not the ability to reason or revise.
Unlike M1 it preserves state across rounds, unlike B1 it removes rather than corrupts semantic
relations, and unlike B2 it retains revision. The contrast between B3 and M4 measures the contribution
of explicit cross-record relations in the otherwise complete workflow. Relative to M4,
B3 lowers mean test by 1.6 and 2.2 percentage points and search AUC by 2.0 and 2.1 for
GPT-5.6 Sol and DeepSeek-V4-Pro, respectively. All reported differences are differences
of run means.

\paragraph{Metrics and Mean Contrasts.}
Let $b_{p,v,0}$ be the validation score of the shared initial Harness, and let $b_{p,v,w}$ be the
best-so-far validation score after all candidates in wave $w$ have been evaluated. For the two-wave
structural-ablation protocol, we use the normalized trapezoidal area
\begin{equation}
    \operatorname{AUC}_p(v)=\frac{1}{W}\sum_{w=1}^{W}
    \frac{b_{p,v,w-1}+b_{p,v,w}}{2}, \qquad W=2.
\end{equation}
This value remains on the validation-score scale and combines the timing and magnitude of
improvements. Wave-level checkpoints prevent the arbitrary completion order of candidates within
a parallel wave from changing the metric.

Let $Y_p(v)$ denote either the repeat-mean validation-selected test or search AUC for variant $v$
under proposer $p$. The main-path effects are
\begin{equation}
    \Delta^{p}_{k}=Y_p(\mathrm{M}k)-Y_p(\mathrm{M}(k-1)), \qquad k\in\{1,2,3,4\}.
\end{equation}
The branch contrasts are $Y_p(\mathrm{B1})-Y_p(\mathrm{M2})$,
$Y_p(\mathrm{B2})-Y_p(\mathrm{M3})$, and $Y_p(\mathrm{B3})-Y_p(\mathrm{M4})$.
The difference between M4 and M0 measures the overall gain of the complete CIG design. Each table cell
reports the absolute metric, while the parenthesized value reports the difference in means from the
corresponding reference.

\paragraph{Interpretation and proposer dependence.}
M2 improving over M1 while B1 falls
below M2 supports the value of correct relation semantics. M3 improving over M2 while B2 falls
below M3 supports revision-aware persistence rather than passive accumulation. M4 improving over
M3 supports direct graph use during selection. Search AUC and test are interpreted separately because a component can
accelerate discovery without changing the final selected Harness, or improve the endpoint without
helping early search. Results are reported separately for GPT-5.6 Sol and DeepSeek-V4-Pro. We
compare the signs and magnitudes of the mean contrasts across proposers.
Each difference measures a component's contribution relative to its specified reference configuration.

\subsection{Run-to-Run Variability of Structural Ablations}
\label{app:stage-three-variance}

Table~\ref{tab:structural-variance} accompanies Table~\ref{tab:structural-ablation}, retaining
the two proposer groups and reporting test means and sample standard deviations across five runs.

\begin{table}[htbp]
    \centering
    \caption{Companion statistics for Table~\ref{tab:structural-ablation}.
    Test scores are mean $\pm$ sample standard deviation in percentage points, grouped by proposer.}
    \label{tab:structural-variance}
    \footnotesize
    \setlength{\tabcolsep}{3pt}
    \renewcommand{\arraystretch}{1.15}
    \begin{tabular}{@{}p{\dimexpr0.42\textwidth-4pt\relax}>{\centering\arraybackslash}p{\dimexpr0.18\textwidth-4pt\relax}>{\centering\arraybackslash}p{\dimexpr0.18\textwidth-4pt\relax}@{}}
        \toprule
        Variant & GPT-5.6 Sol & DeepSeek-V4-Pro \\
        \midrule
        \TableFourVarianceRows
        \bottomrule
    \end{tabular}
\end{table}

\subsection{Selection Eligibility and Verdict Semantics}
\label{app:verdict-semantics}

Candidate legality, intervention fidelity, and hypothesis support answer different questions.
Legality concerns executable artifacts, allowed dependencies, and resource contracts. Fidelity
asks whether the proposed modification was realized. Support asks whether the measured effects
agree with the prediction. Final selection uses eligible measured performance, independently of whether
the candidate's motivating explanation is supported.

For score-based observations, the numerical outcome layer first checks improvement over the prior
incumbent and assigns \texttt{globally\_improving}. Otherwise it compares against the best scored
parent and assigns \texttt{locally\_supported} for an increase, \texttt{refuted} for a decrease,
or \texttt{uncertain} otherwise. These labels summarize measured score changes.
The reasoning layer reads these labels together with the measured
effects and registered predictions to revise the current hypothesis assessment. Revision
changes the interpretation available to later calls, not the recorded scores or their provenance.
Task-level scores support task-level assessments. Sample-level transitions and execution-fidelity
judgments are included only when the corresponding evidence is available.

At the hypothesis level, support means that the measured result agrees with the stated prediction
over its specified scope. Local support applies when only a narrower task or condition agrees.
Refutation requires an observed violation of the stated falsification condition. Inconclusive
or unavailable evidence leaves the assessment uncertain. Scope refinement retains the original
claim and records the narrower conditions under which it remains consistent with observations.
Agreement in score alone does not establish why an intervention worked.

A failed aggregate target therefore refutes that target for the tested implementation and
setting, not every constituent mechanism. Task-level scores can resolve directional trade-offs.
Sample-level corrected/regressed claims require paired traces and cannot be inferred from
aggregate scores. Infrastructure failures remain distinct from measured Harness failures.

\subsection{Operation Interfaces and Revision Details}
\label{app:operation-details}

The operational interface separates factual observations, revisable assessments, and selection
decisions. The role specifications below describe the inputs, reasoning operations, and
outputs of each stage.

\paragraph{Observe and reassess.}
Inputs are the current graph, permitted evaluation history, and predictions linked to the
evaluated interventions. The permitted history includes candidate artifacts, validation scores,
reports, and available execution traces. The observer accesses these records through source
references as needed, within the configured context and computation limits.
The output identifies grounded evidence, current hypothesis assessments,
unresolved questions, and conflicts. Reassessment compares the predicted effects with the
available observations, distinguishes a failed joint target from a partially supported local
prediction, and preserves references to the relevant experiment. Existing observations remain
unchanged when their interpretation is revised.

\paragraph{Plan and select.}
Plans name their parent Harnesses, supporting evidence, proposed changes, predicted effects,
and falsification conditions. Each prediction specifies the expected direction and relevant
task or behavior, while its falsification condition states what observed result would count
against it. In the Wide setting, the planner generates eight plans and the selector chooses
four per wave. Planning and candidate evaluation are separate, so only the four implemented
plans consume candidate evaluations.
Selection reads the plans together with the current relational
state. It uses the seven criteria described in Section~\ref{sec:method} to compare plans and
portfolio redundancy, including the relevance of each plan to unresolved or revised hypotheses.
Redundancy concerns overlapping target mechanisms and expected effects, while complementarity
concerns distinct failure modes, competing explanations, or alternative mechanisms.
The portfolio contains an evidence-grounded refinement and a branch or restart and does not
allocate every slot to one mechanism family. Selection is a proposer judgment under these
constraints, not a deterministic maximization of a weighted score. It returns selected plans
without rewriting them, together with reasons for selection and rejection.

\paragraph{Commit and carry forward.}
The evaluated candidate, parent-relative effects, and numerical verdict are retained as an
experimental record. Their dependencies connect the result to its motivating prediction.
Comparison first separates whether the intended change was executed, whether the measured
direction agrees with the prediction, and whether the evidence supports the proposed explanation.
Missing execution evidence is not inferred from a score increase.
The revised assessment forms part of the state read by subsequent observation, planning, and
selection. Append-only persistence retains the experimental records without revising earlier
assessments, whereas revision-aware persistence allows the interpretation guiding later decisions
to change as evidence accumulates.

\paragraph{Assessment example.}
In Appendix~\ref{app:case-trace}, the prediction is to recover LawBench without losing the
other gains and to exceed $58.00$ aggregate performance. The recorded outcome is LawBench
$+6.00$, USPTO $-6.67$, unchanged Symptom2Disease, and aggregate $-0.22$. Its numerical label
is \texttt{refuted}. At the assessment level, the joint target failed, the local LawBench
prediction remains consistent with the observation, and the coverage explanation remains
unresolved. This illustrates the distinction between a numerical verdict and the scoped
interpretation carried by revision.

\subsubsection{Prompt Excerpts}
\label{app:prompt-excerpts}

The following prompt excerpts illustrate the method's role boundaries and core reasoning
operations. The final box describes the falsifiable-hypothesis instructions for Full History.

\begin{tcolorbox}[enhanced,breakable,colback=black!2,colframe=black!30,
boxrule=0.5pt,arc=1pt,left=7pt,right=7pt,top=6pt,bottom=6pt,
fonttitle=\bfseries,title={Role-Specific Instructions}]
\small\raggedright
\textbf{Shared context and execution boundaries}\par\smallskip Use the current graph state and its source evidence to carry out the assigned reasoning operation. Consult the underlying records when a summary is insufficient, and keep measured facts distinct from their interpretations. Perform only the assigned role. Do not collapse observation, planning, selection, and implementation into a single step. The runner alone commits graph transitions and measured outcomes, updates the frontier, and manages held-out testing. Never invent or modify runner-grounded validation facts.
\par\medskip
\textbf{Observer}\par\smallskip Separate high-confidence observations from inferred explanations. Preserve contradictory and low-frequency evidence. Do not design, rank, select, or implement candidates.
\par\medskip
\textbf{Selector}\par\smallskip Select a portfolio, not merely the highest individual scores.
\par\medskip
\textbf{Implementer}\par\smallskip The plan fixes the experimental question and primary intervention, but you retain freedom to choose the strongest faithful implementation.
\end{tcolorbox}

\begin{tcolorbox}[enhanced,breakable,colback=black!2,colframe=black!30,
boxrule=0.5pt,arc=1pt,left=7pt,right=7pt,top=6pt,bottom=6pt,
fonttitle=\bfseries,title={Hypothesis Revision}]
\small\raggedright
Read the current hypothesis records together with their linked evidence, interventions, and newly available outcomes. For each existing hypothesis affected by a new outcome:
\par\smallskip
1. Reference its existing hypothesis ID. Preserve its original claim and    previous assessment rather than replacing the historical record.
\par\smallskip
2. Compare the stated prediction with the observed outcome. Distinguish    intervention fidelity, local effects, and the overall performance objective.
\par\smallskip
3. Propose a current assessment of supported, contradicted, or unresolved    at the scope justified by the evidence. A failed joint objective does not    by itself refute every local prediction or the broader explanation.
\par\smallskip
4. State any justified scope refinement and cite the supporting and conflicting    evidence. Preserve uncertainty when the evidence cannot discriminate.
\par\smallskip
5. Explain the implications for subsequent experiments without generating,    ranking, selecting, or implementing candidate plans.
\par\medskip
Return a revision proposal containing hypothesis\_id, previous\_assessment, proposed\_assessment, previous\_scope, proposed\_scope, evidence\_ids, rationale, and unresolved\_questions. Create a new hypothesis ID only for a substantively new claim and identify its relationship to the earlier claim. Do not alter measured outcomes. The runner validates references and commits revisions while preserving earlier versions and their provenance.
\end{tcolorbox}

\begin{tcolorbox}[enhanced,breakable,colback=black!2,colframe=black!30,
boxrule=0.5pt,arc=1pt,left=7pt,right=7pt,top=6pt,bottom=6pt,
fonttitle=\bfseries,title={Falsifiable Hypotheses (Full History)}]
\small\raggedright
Formulate three falsifiable hypotheses targeting different mechanisms. For each hypothesis:
\par\smallskip
1. Cite the observed evidence in the available history. Distinguish measured    facts from inferred explanations and retain conflicting observations.
\par\smallskip
2. State the explanatory claim, its applicable scope, and its uncertainty.
\par\smallskip
3. Identify the parent Harness and one primary intervention. State which    mechanisms should remain unchanged.
\par\smallskip
4. Predict an observable effect and specify what result would contradict    the prediction. Distinguish local predictions from the overall objective.
\par\smallskip
5. Explain why the experiment is informative given previous attempts.    Revisit a failed direction only with a stated change or new evidence.
\par\medskip
Do not invent measurements or treat an improved score as proof of an explanation. Use the available history to choose and implement the next candidates. Prototype the proposed mechanism before writing the final system.
\end{tcolorbox}

\subsection{Proposer Token Usage}
\label{app:token-usage}

Beyond candidate-evaluation counts, we examine how proposer computation translates into
Harness quality. AutoSaddler~\citep{park2026autosaddler} relates higher optimizer-side cost
per patch to improved search outcomes and more selective evaluation.
Vesper~\citep{ishibashi2026vesper} shows that investing more computation per candidate can
improve discovery under a common token budget. Following this budget-based perspective,
we examine CIG with GPT-5.4 as proposer and Qwen3.7-Flash as solver on USPTO,
Symptom2Disease, and LawBench.
Table~\ref{tab:token-usage} compares complete Full History and CIG Wide searches, each
evaluating eight candidates, and a CIG checkpoint within Full History's proposer token
budget. The checkpoint is reconstructed from the search trajectory and includes all
proposer operations and retries incurred up to that point. Among candidates with completed
evaluations at the checkpoint, validation selects the Harness whose test score is reported.
Proposer usage covers input and output tokens across discovery operations, separately from
solver execution during candidate evaluation.

\begin{table}[htbp]
    \centering
    \caption{Proposer token usage and validation-selected test performance in a representative
    configuration. Token ratios use the complete Full History run as reference.}
    \label{tab:token-usage}
    \small
    \setlength{\tabcolsep}{5pt}
    \begin{tabular}{@{}lrrrr@{}}
        \toprule
        Method / checkpoint & Tokens (M) & Ratio & Candidates & Test (\%) \\
        \midrule
        Full History (complete) & 0.530 & $1.00\times$ & 8 & 41.3 \\
        CIG Wide (budget checkpoint) & 0.530 & $1.00\times$ & 5 & 46.8 \\
        CIG Wide (complete) & 0.820 & $1.55\times$ & 8 & 52.9 \\
        \bottomrule
    \end{tabular}
\end{table}

At the common budget of 0.530M proposer tokens, CIG reaches 46.8\% test with five evaluated
candidates, exceeding Full History's complete eight-candidate search by 5.5 percentage points.
Continuing CIG to eight candidates raises test to 52.9\% at 1.55 times the proposer token
usage, a gain of 11.6 points over Full History. Thus, in this configuration, CIG achieves
higher Harness quality both at the common proposer budget and with additional computation.
This comparison highlights the importance of how proposer computation is organized, alongside
how much is used. CIG connects locally scoped operations through persistent improvement state,
so prior experimental dependencies and updated assessments guide subsequent proposals.
The checkpoint and complete-run results support this workflow's effective use of a common
proposer budget and its ability to obtain further gains with additional computation in this setting.

\subsection{Experimental Configuration and Reporting}
\label{app:configuration}

\paragraph{Code availability.}
We will release the code publicly upon acceptance.

Unless otherwise stated, our experimental results are means over five independent runs.
All entries in Table~\ref{tab:main-benchmark-results}, including our runs and the published
baselines, are means over three runs. Within each controlled comparison, methods use the same
task solver, data partitions, and initial Harness artifacts. Initialization is fixed before
search and does not use held-out test feedback.
Figure~\ref{fig:search-dynamics} shows individual searches on LawBench with GPT-5.4 as proposer
and DeepSeek-V4-Flash as solver. Search evaluates each candidate once. For the figure, each
Harness is re-evaluated five times on the fixed 50-task validation set, and the curve uses
these means over candidates available at each completed wave. These post-search measurements
do not guide discovery or candidate selection. The case studies also describe individual searches.
Tables~\ref{tab:proposer-dependence}, \ref{tab:structural-ablation}, and
\ref{tab:structural-variance} use the SWE-bench Verified 100/50/100 train/validation/test
partition in Table~\ref{tab:data-partitions}, with a binary success result
per task in each run, then report the mean across five runs. Delta columns subtract the
corresponding run means.

\paragraph{Model configurations.}
The agent-benchmark comparison uses GPT-5 mini for both proposal and task execution.
The solver study fixes GPT-5.4 as proposer and varies Qwen3.5-Plus, Qwen3.7-Flash,
DeepSeek-V4-Flash, GLM-5.2-Fast-Preview, and MiniMax-M2.5.
The proposer study fixes Qwen3.5-Plus as solver and varies GPT-5.5, GPT-5.6 Sol,
GLM-5.3, Kimi K3, and DeepSeek-V4-Pro. Structural ablations use Qwen3.7-Flash as solver
with GPT-5.6 Sol or DeepSeek-V4-Pro as proposer. Each model pairing remains fixed throughout
its search and final evaluation.

\paragraph{Data partitions.}
Table~\ref{tab:data-partitions} lists the classification partitions and the agent-benchmark
partitions used for the HarnessFix-aligned comparison. Training examples are available for
Harness construction, validation guides discovery and selection, and test is reserved for
final evaluation. The Terminal-Bench partition is reconstructed following the reference
protocol using the public split sampler with seed 20260522.

\paragraph{Published comparison results.}
The published block of Table~\ref{tab:main-benchmark-results} uses the results reported by
HarnessFix. Our block uses the benchmark environments and evaluation setup described by that
protocol, including the benchmark-specific initial Harnesses listed below. The two blocks
report the same task-completion metric and three-run mean. Within our block, CIG and
Full History share the task partitions and initialization.

\begin{table}[htbp]
    \centering
    \caption{Dataset partition sizes. Classification rows count examples, and agent-benchmark
    rows count tasks.}
    \label{tab:data-partitions}
    \small
    \begin{tabular}{lrrr}
        \toprule
        Benchmark & Train & Validation & Test \\
        \midrule
        USPTO-50k & 50 & 30 & 100 \\
        Symptom2Disease & 200 & 50 & 212 \\
        LawBench & 200 & 50 & 100 \\
        \midrule
        SWE-bench Verified (HarnessFix protocol) & 100 & 50 & 100 \\
        AppWorld & 90 & 45 & 90 \\
        Terminal-Bench 2.0 Verified & 34 & 17 & 34 \\
        \bottomrule
    \end{tabular}
\end{table}

\paragraph{Initialization and search schedule.}
Classification search starts from the same no-memory and all-training-example few-shot
Harnesses. Their frozen validation results are reused within a matched solver configuration,
rather than reevaluated for every search method. Agent-benchmark methods start from the same
benchmark-specific initial Harness: mini-swe-agent for SWE-bench Verified, simplified ReAct
for AppWorld, and Terminus-2 for Terminal-Bench.
The controlled solver, proposer, and structural studies use eight candidate evaluations.
Full History and CIG (Deep) use four waves of two candidates, while CIG (Wide) uses two waves
of four. The structural Full History control instead shares CIG's Wide schedule and staged
calls. In CIG (Wide), each wave proposes eight plans and evaluates four selected candidates.

Results use the validation-selected Harness. Selection regret measures the gap to the test
oracle within each method's generated candidate pool. The final Harness selection rule is
distinct from plan selection during discovery.

\subsection{Detailed Case Study}
\label{app:case-trace}

This case follows a two-wave CIG (Wide) search with GPT-5.4 as proposer and Qwen3.5-Plus
as solver on USPTO-50k, Symptom2Disease, and LawBench. It shows how an apparently beneficial
candidate becomes the starting point for a more specific experimental question. The excerpts
below condense the recorded nodes without changing their predictions, intervention, or outcomes.
All scores are discovery-time validation scores from this single run.

\begin{center}
$\textcolor{evidencebar}{E_2}
\xrightarrow{\text{grounds}}\textcolor{hypothesisbar}{H_2}
\xrightarrow{\text{tested by}}\textcolor{interventionbar}{I_2}
\xrightarrow{\text{yields}}\textcolor{outcomebar}{O_2}
\xrightarrow{\text{informs reassessment}}\textcolor{hypothesisbar}{H_2}$
\end{center}

\begin{casenodebox}{evidencebg}{evidencebar}{$E_2$ | Evidence: expose a trade-off within an improvement}
\small
\textbf{Source comparison.} The first wave replaces \texttt{fewshot\_all} with
\texttt{similarity\_retrieval\_fewshot}, selecting demonstrations by query-conditioned
lexical similarity.

\medskip
\textbf{Recorded observation.} Aggregate performance rises from $56.90$ to $58.00$.
USPTO improves from $36.67$ to $40.00$ and Symptom2Disease from $74.00$ to $88.00$,
while LawBench falls from $60.00$ to $46.00$.

\medskip
\textbf{Implication for the next decision.} Retrieval is promising but its benefit is not
uniform. The linked task-level evidence identifies a specific regression to investigate
without discarding the gains of the parent Harness.
\end{casenodebox}

\begin{casenodebox}{hypothesisbg}{hypothesisbar}{$H_2$ | Hypothesis: turn the trade-off into a testable prediction}
\small
\textbf{Proposed explanation.} Query-conditioned retrieval removes irrelevant demonstrations,
but selecting only by similarity may remove the broader coverage needed by LawBench.
The recorded uncertainty is medium.

\medskip
\textbf{Prediction registered before evaluation.} A coverage-preserving refinement should
recover LawBench, retain the gains on the other tasks, and improve aggregate performance
beyond the parent's $58.00$.

\medskip
\textbf{Failure condition.} No LawBench recovery, or offsetting losses that leave aggregate
performance at or below $58.00$, would defeat the proposed joint improvement target.
The hypothesis therefore specifies what the next experiment should establish, rather than
merely requesting a higher-scoring candidate.
\end{casenodebox}

\begin{casenodebox}{interventionbg}{interventionbar}{$I_2$ | Intervention: test coverage while retaining the retrieval backbone}
\small
\textbf{Parent and candidate.} Refine \texttt{similarity\_retrieval\_fewshot} into
\texttt{similarity\_coverage\_backfill\_fewshot}.

\medskip
\textbf{Executable change.} Allocate $60\%$ of the prompt budget to nearest lexical neighbors,
then fill the remaining slots using answer-space and token-coverage novelty.

\medskip
\textbf{Preserved constraints.} Keep raw exemplar prompting and query-conditioned lexical
retrieval, with one prediction-time solver call, no train-time solver calls, and linear storage.
This makes the proposed remedy concrete and connects the evaluated code change to the
coverage explanation, rather than replacing the entire Harness with an unrelated design.
\end{casenodebox}

\begin{casenodebox}{outcomebg}{outcomebar}{$O_2$ | Outcome: retain the local recovery and the failed joint target}
\small
\textbf{Measured effects relative to the parent.}
\begin{center}
\begin{tabular}{lrrr}
\toprule
Metric & Parent & Intervention & Change \\
\midrule
USPTO & $40.00$ & $33.33$ & $-6.67$ \\
Symptom2Disease & $88.00$ & $88.00$ & $0.00$ \\
LawBench & $46.00$ & $52.00$ & $+6.00$ \\
Aggregate & $58.00$ & $57.78$ & $-0.22$ \\
\bottomrule
\end{tabular}
\end{center}
\textbf{Stored verdict.} The record labels the outcome \texttt{refuted} and commits it to
$G_2$ as a negative but informative result. The aggregate target was not met.

\medskip
\textbf{Scope of the evidence.} LawBench's recovery supports the local directional
prediction, while USPTO's regression violates the joint no-regression goal.
The broader coverage explanation remains unresolved.
The graph retains the measured effects alongside the verdict and their links to the
tested hypothesis, intervention, and parent.
\end{casenodebox}

\paragraph{What the graph makes explicit.}
The chain preserves an experimental argument, not just a ranking of Harnesses.
$E_2\!\rightarrow\!H_2$ links the proposed explanation to the observation that motivated it.
$H_2\!\rightarrow\!I_2$ connects a prospective prediction to an executable change.
$I_2\!\rightarrow\!O_2$ records what happened when that change was tested.
The outcome-to-hypothesis link keeps the negative verdict attached to its evidential scope:
the specific joint prediction failed even though one intended effect appeared.
These relations make it possible to inspect why the experiment was proposed and which parts
of its rationale remain unresolved, rather than reconstructing that argument from a score
and a code diff.

First-wave evidence informs the second-wave coverage intervention in the trace above.
A separate successful search provides the updated-assessment branch for the comparison below.

\subsubsection{State Intervention: Updated versus Retained Assessment}
\label{app:case-state-intervention}

The updated branch comes from a four-wave CIG search with GPT-5.4 as proposer and
Qwen3.5-Plus as solver. Each wave proposes eight plans and evaluates two candidates.
The comparison starts after the first-wave top-$k$ retrieval candidate is evaluated on
USPTO, Symptom2Disease, and LawBench. Its validation scores are $40.00$, $84.00$,
and $48.00$, respectively, with an equal-weight mean of $57.33$.

\paragraph{Assessment contrast.}
The earlier explanation concerns dilution from including all demonstrations.
The updated explanation identifies a more specific trade-off: local retrieval may remove
the broader demonstration coverage needed by LawBench, whose score fell from $60$ to $48$.
These are summaries of the recorded hypotheses. The refined explanation is recorded under
a new hypothesis identifier. Both branches start from the same post-evaluation checkpoint
with identical new outcomes, raw history, models, and next-wave plan and evaluation counts,
differing only in the supplied assessment state.

\begin{tcolorbox}[enhanced,breakable,colback=black!2,colframe=black!30,
fonttitle=\bfseries,title={Assessment $\rightarrow$ plan $\rightarrow$ Harness $\rightarrow$ outcome}]
\small
The two assessments motivate distinct directions for the next intervention.
\par\smallskip
\renewcommand{\arraystretch}{1.14}
\begin{tabularx}{\linewidth}{@{}>{\raggedright\arraybackslash}p{0.17\linewidth}YY@{}}
\toprule
 & \textbf{Updated assessment} & \textbf{Retained assessment} \\
\bottomrule
\end{tabularx}
\par\vspace{1pt}
\begin{tabularx}{\linewidth}{@{}>{\raggedright\arraybackslash}p{0.17\linewidth}YY@{}}
\textbf{Selected plan and rationale}
& Add coverage anchors to local retrieval. The selector cites the LawBench decline
from $60$ to $48$ and targets recovery while preserving the benefits of query-conditioned retrieval.
& Refine local retrieval. The aggregate gain supports prioritizing query-relevant
demonstrations, motivating more precise selection to improve performance further. \\
\end{tabularx}
\par\vspace{1pt}\hrule\vspace{1pt}
\begin{tabularx}{\linewidth}{@{}>{\raggedright\arraybackslash}p{0.17\linewidth}YY@{}}
\textbf{Harness modification}
& Refine the top-$k$ retrieval Harness with a global label/pattern anchor bank alongside
local demonstrations, retaining one prediction-time solver call.
& Expand the retrieval pool and rerank examples using complementary lexical signals
before retaining the top-$k$. Preserve the demonstration budget, prompt format,
and single prediction-time solver call. \\
\end{tabularx}
\par\vspace{1pt}\hrule\vspace{1pt}
\begin{tabularx}{\linewidth}{@{}>{\raggedright\arraybackslash}p{0.17\linewidth}YY@{}}
\textbf{Evaluation result}
& USPTO: $43.33$ ($+3.33$).\newline
Symptom2Disease: $86.00$ ($+2.00$).\newline
LawBench: $52.00$ ($+4.00$).\newline
Macro mean: $\mathbf{60.44}$ ($\mathbf{+3.11}$).
& USPTO: $40.00$ ($0.00$).\newline
Symptom2Disease: $84.00$ ($0.00$).\newline
LawBench: $48.00$ ($0.00$).\newline
Macro mean: $\mathbf{57.33}$ ($\mathbf{0.00}$). \\
\bottomrule
\end{tabularx}
\par\smallskip
Scores are validation percentages. Parenthesized changes are relative to the shared parent.
The macro mean averages the three tasks within this run.
\end{tcolorbox}

The retained assessment directs finer local relevance ranking, whereas the updated assessment
directs complementary demonstration coverage. The resulting Harnesses score $57.33$ and
$60.44$, a $3.11$-point difference that links a changed interpretation to a different
intervention and higher measured performance.

\end{document}